\documentclass[10pt,a4paper]{article}
\usepackage{hyperref}
\usepackage{amsmath,amsfonts}
\usepackage{graphicx} 

\usepackage{verbatim}
\usepackage{enumitem}
\usepackage[width=15cm,height=22cm]{geometry}
\usepackage{authblk}

\newtheorem{hypothesis}{Assumption}

\newtheorem{proposition}{Proposition}[section]
\newtheorem{definition}[proposition]{Definition}
\newtheorem{lemma}[proposition]{Lemma}
\newtheorem{remark}[proposition]{Remark}
\newtheorem{theorem}{Theorem}
\newcommand{\bpf} {\noindent{\sc Proof} : }
\newcommand{\epf} {\hfill$\square$\vspace{.5cm}}

\newcommand{\g} {\Gamma_{m,Z,\lambda}}

\newcommand{\N}{\mathbb{N}}
\newcommand{\R}{\mathbb{R}}

\newcommand{\h}{\mathcal{H}}
\renewcommand{\H}{\mathbf{H}}

\newcommand {\x} {\mathcal{X}}
\newcommand {\y} {\mathcal{Y}}

\newcommand {\zi} {{Z^i}}
\renewcommand {\l} {\ell}
\newcommand{\z}{\mathcal{Z}}
\newcommand{\A}{\mathcal{A}}
\newcommand{\B}{\mathcal{B}}
\renewcommand{\a}{\mathbf{A}}
\renewcommand{\b}{\mathbf{B}}
\DeclareMathOperator*{\argmin}{arg\,min} 

\title{
Equivalence of Learning Algorithms
}

\author[1]{Julien Audiffren}
\author[2]{Hachem Kadri}
\affil[1]{CMLA, ENS Cachan,Cachan,France}
\affil[2]{QARMA, Aix-Marseille Université, CNRS, LIF, Marseille, France}
\date{}

\begin{document}

\maketitle


\begin{abstract}


The purpose of this paper is to introduce a concept of equivalence between machine learning algorithms.
 We define two notions of algorithmic equivalence, namely, weak and strong equivalence. These notions are of paramount importance for identifying when learning properties from one learning algorithm can be transferred to another. Using regularized kernel machines as a case study,  we illustrate the importance of the {introduced} equivalence concept by analyzing the relation between kernel ridge regression~(KRR) and $m$-power regularized least squares regression~(M-RLSR) algorithms.

\end{abstract}

\section{Introduction}



Equivalence is a fundamental concept that defines a relationship between two objects and allows the inference of properties of one object from the properties of the other. In the field of machine learning, several theoretical concepts have been proposed in order to estimate the accuracy of learning algorithms, among which complexity of the hypothesis set~\cite{Vapnik1991, EVG00, Bartlett02}, stability~\cite{Bousquet} and robustness~\cite{Xu10}, but little can be said concerning how these learning properties can be moved from one learning algorithm to another one.  This raises the question of how equivalence between two learning algorithms might be defined such that some learning characteristics could be transferred between them. 

When are two learning algorithms equivalent? More precisely, given $\{Z,\lambda_i,\mathcal{A}_i\}$, $i=1,2$, where $Z$ is a training set, $\mathcal{A}_i$ is a learning algorithm that constructs to every training set $Z$ a decision function $f^i_{Z,\lambda}$ and $\lambda_i$ is a tuning parameter that balances the trade-off between fitness of $f^i_{Z,\lambda}$ to the data $Z$ and smoothness of $f^i_{Z,\lambda}$, under what conditions is $\mathcal{A}_1$ equivalent to $\mathcal{A}_2$? 
%
Many learning algorithms can be formulated as an optimization problem. In this case, it is often thought that two learning algorithms are equivalent if their associated optimization problems are. {One purpose of this paper is to point out that showing that two optimization problems are equivalent is not adequate evidence that the underlying learning algorithms are exchangeable with each other or even share some learning properties}. Indeed, restricting the analysis of equivalence between learning algorithms to that of their optimization problems tends to omit details and ignore steps forming  the whole learning {mechanism} such as the selection of the tuning parameter $\lambda$ or the change of the learning set $Z$. The notion of equivalence between learning algorithms needs thus to be clearly defined.

The concept of \textit{algorithmic equivalence} in machine learning has never been properly defined; the degree of difference between an optimization problem and its associated learning algorithm is not defined anywhere, nor has an exact definition. In this work, we rigorously define two notions of equivalence for learning algorithms and show how to use these notions to transfer stability property from a learning algorithm to another. Algorithmic stability is of particular interest since it provides a
 sufficient condition
for a learning algorithm to be consistent and generalizing~\cite{Shalev09}.
The first notion of equivalence we define, called weak-equivalence, is related in some way to the equivalence between the associated minimization problems. By weak equivalence, we would like to emphasize here that 
{this equivalence holds only when the algorithms are evaluated on a given training set $Z$}. 
This matches in some manner the equivalence between the optimization problems since the objective function to minimize is evaluated only for the set $Z$ of training examples ${(x_i,y_i)}_{i=1}^n$, which is fixed in advance. The second notion is  stronger, in the sense that two learning algorithms are strongly equivalent when their equivalence does not depend on the training set $Z$.


%
%

As a case study, we consider regularized kernel methods which are learning algorithms with a regularization over a reproducing kernel Hilbert Space~(RKHS) of functions. In particular, we study the regularized least squares regression problem when the RKHS regularization is raised to the power of $m$~\cite{Mendelson10,Steinwart09}, where $m$ is a variable real exponent, and design an efficient algorithm for computing the solution, called \mbox{M-RLSR}~(\mbox{$m$-power} regularized least squares regression).  Using our algorithmic equivalence concept, we analyze the relation between \mbox{M-RLSR} and kernel ridge regression~(KRR) algorithms.


In this paper, we make the following contributions: \textbf{1)} we formalize the concept of equivalence between two learning algorithms and define two notions of algorithmic equivalence, namely, weak and strong equivalence~(Section~\ref{weak}). \textbf{2)} We show that the weak equivalence is not sufficient to allow the transfer of learning properties, such as stability, while strong equivalence is. Moreover, we provide sufficient assumptions  under which the transfer of stability still holds even in  the weak equivalence case~(Section~\ref{transfer}). \textbf{3)} As a case study, we consider the equivalence between KRR and M-RLSR. More precisely,  we derive a semi-analytic solution to the M-RLSR optimization problem, we design an efficient algorithm for computing it, and we show that M-RLSR and KRR algorithms are weakly  and not strongly equivalent~(Section~\ref{MRLSR}).

\section{Notations and Backround}
\label{NotBack}

In the following, $\x$ will denote the input space, $\y$ the output space, $ \z^n=(\x \times \y)^n$, $\z= \cup_{n\ge 1} \z^n$, $Z \in \z$ a training set, $\h\subset \y^\x$ the Banach space of hypotheses (for instance a separable reproducing kernel Hilbert space~(RKHS)). Here and throughout the paper we use both notations $\vert Z  \vert$ and $\#(Z)$ to denote the cardinal of the set $Z$.

Following the work of Bousquet and Elisseeff~\cite{Bousquet}, a learning algorithm is defined as a mapping that takes a learning set made of input-output pairs and produces a function $f$ that relates inputs to the corresponding outputs. For a large class of learning algorithms, the function $f$ is obtained by solving an optimization problem. So before talking about \textit{algorithmic equivalence}, it is helpful to discuss equivalence between optimization problems. A mathematical optimization problem~\cite{boyd}  has the form
\begin{align*}
minimize \quad & l(f)\\
subject\  to \quad & l_i(f) \leq b_i, \quad i=1,\ldots,q.
\end{align*}
Here $f$ is the optimization variable of the problem, the function $l: \mathcal{H}\rightarrow \mathbb{R}$ is the objective function, the functions $l_i$ are the (inequality) constraint functions, and the constants $b_i$ are the limits for the constraints. A function $f^*$ is called optimal, or solution of the problem, if it has the smallest objective value among all functions that satisfy the constraints. We consider here only strictly convex objective and constraint functions, so that the minimization problem have an unique solution and $f^*$ is well defined.
Two optimization problems are equivalent if both provide the same optimal solution.

The machine learning literature contains studies showing equivalence between learning algorithm~(see, e.g.,~\cite{Giroso1998, Rudin09, Jaggi13}); however, most of them have focused only on the equivalence that may occur between the associated optimization problems. In this sense, an equivalence between two optimization problems offers a way to relate the associated learning algorithms. However, this is not sufficient to decide whether the optimization equivalence allows to transfer theoretical properties from one learning algorithm to the other. 
%
%
%
%
%
%
%
%
A work that have indirectly supported this view is that of Rifkin~\cite{rifkin}, who studied learning algorithms related to Thikonov and Ivanov regularized optimization.
%
%
%
In~\cite[Chapter~5]{rifkin}, it was shown that even though these optimization problems are equivalent, i.e., they give the same optimal solution, the associated learning algorithms have not the same stability properties.
From this point of view, equivalence between minimization problems does not imply that the underlying algorithms share the same learning characteristics. This consideration is closely related to our goal of 
identifying when properties of a learning algorithm can be moved from one to another. The concept of \textit{algorithmic equivalence} emerges in response to this question.




\section{Weak and Strong Equivalence Between Learning Algorithms}\label{weak}

In this section, we provide a rigorous definition of the concept of equivalence between machine learning algorithms. The idea here is to extend the notion of equivalence of optimization problems to learning algorithms. 
%
We first start by recalling the definition of a learning algorithm as given by Bousquet and Elisseeff~\cite{Bousquet}, and for simplicity we restrict ourselves to learning algorithms associated to strictly convex optimization problems.

\begin{definition}(Learning Algorithm).
\label{def_algo}
A learning algorithm $\A$ is a function \mbox{$\A: \z \rightarrow \h$} which maps learning set $Z$ onto a function A(Z), such that 
\begin{equation}\label{eq def algo}
\A(Z) = \argmin_{g \in \h} R(Z,g),
\end{equation}
where $R(Z,\cdot)$ is a strictly convex objective function. 
\end{definition}

Since we consider only strictly convex objective functions, the minimization problem has an unique solution and \eqref{eq def algo} is well defined. From this definition, the following definition of equivalence between algorithm naturally follows.

\begin{definition} \label{def equiv}(Equivalence).
Let $Z$ be a training set. Two algorithms $\A$ and $\B$ are equivalent on $Z$ if and only if
$\A(Z)=\B(Z)$.
\end{definition}

In other words, let $\A$~(resp. $\B)$ be a learning algorithm associated to the optimization problem $R$~(resp. $S$), then $\A$ and $\B$ are equivalent on $Z$ if and only if the optimal solution of $R(Z,\cdot)$ is the optimal solution of $S(Z,\cdot)$.
It is important to point out that the optimal solutions of $R$ and $S$ are computed for a set $Z$, and even though they are equal on $Z$,  there is no guarantee that this remains true if $Z$ varies. This means that the two algorithms $\A$ and $\B$ provide the same output with the set $Z$, but this may not be necessarily the case with another set $Z'$.

In this paper, we pay special attention to regularized learning algorithms. These algorithms depend on a regularization parameter that plays a crucial role in controlling the trade-off between overfitting and underfitting.
It is important to note that many widely used regularized learning algorithms are families of learning algorithms. Indeed, each value of the regularization parameter $\lambda$ defines a different minimization problem. As an example, we consider Kernel ridge regression~\cite{saunders98} which is defined as follows%
\begin{equation}\label{KRR}
\begin{aligned}
&\argmin_{f \in \h} \frac{1}{\vert Z \vert}\sum_{(x,y) \in Z} (y - f(x))^2 + \lambda \|f \|^2_\h.
\end{aligned}
\end{equation}
In the following, we denote by $\mathbf{A}(\cdot)$ a regularized learning algorithm indexed by a regularization parameter in $ \R^*_+$. In other words, $\forall \lambda \in \R^*_+,$ $\A(\cdot)\doteq \mathbf{A}(\lambda)(\cdot)=\mathbf{A}(\lambda,\cdot)$ defines a learning algorithm as in definition \ref{def_algo}. 
Note that all the results we will state can be naturally extended to a larger class of family of learning algorithms.

\textbf{Equivalence of regularized learning algorithms.} We now define the notion of weak equivalence between regularized learning algorithms as an extension of definition \ref{def equiv}. 
To illustrate this equivalence, we provide some basic examples in this section and an in-depth case study  in Section~\ref{MRLSR}.

\begin{definition}\label{weak equivalence}(Weak Equivalence).
Let $\mathbf{A}(\cdot)$ and $\mathbf{B}(\cdot)$ two regularized learning algorithms on $\R^*_+ \times \z$. Then $\mathbf{A}(\cdot)$ and $\mathbf{B}(\cdot)$ are said to be weakly equivalent if and only if 
$\exists \Phi_{\a \rightarrow \b} : \R^*_+ \times \z \mapsto   \R^*_+$ such that
\begin{enumerate}
\item $\forall Z\in \z$, $\Phi_{\a \rightarrow \b}(\cdot,Z)$ is a bijection from $\R^*_+$ into $\R^*_+,$
\item $\forall Z \subset \z,\quad \forall \lambda \in \R^*_+,$  $\mathbf{A}(\lambda)$ and $\mathbf{B}(\Phi_{\a \rightarrow \b}(\lambda,Z))$ are equivalent on $Z$. 
\end{enumerate}
\end{definition}

In the particular case where $\Phi_{\a \rightarrow \b}$ does not depend on $Z$, this assertion becomes much stronger than the weak equivalence, and will be referred as strong equivalence.

\begin{definition}\label{strong equivalence}(Strong Equivalence).
Let $\mathbf{A}(\cdot)$ and $\mathbf{B}(\cdot)$ two regularized learning algorithms on $\R^*_+ \times \z$. Then $\mathbf{A}(\cdot)$ and $\mathbf{B}(\cdot)$ are said to be strongly equivalent if and only if it exists
$\Phi_{\a \rightarrow \b}$, a bijection from $\R^*_+$ into $\R^*_+$ such that $ \mathbf{A}(\cdot)=\mathbf{B}(\Phi_{\a \rightarrow \b}(\cdot))
$
where the equality is among functions from $\R^*_+$ into $\h^\z$.
\end{definition}

This notion of weak equivalence is frequently encountered in machine learning algorithms. For instance, it naturally occurs when using Lagrangian duality and when transiting from Ivanov's to Thikonov's method~\cite[Chapter~5]{rifkin}.
%
%
Note that weak and strong equivalence between two learning algorithms have some immediate implications for interpreting their regularization paths (see the supplementary material for more details).
The natural question which now arises is whether by knowing some learning properties of $\mathbf{A}$ and the weak equivalence of $\mathbf{A}$ and $\mathbf{B}$ it is possible to deduce learning properties for $\mathbf{B}$. This question is studied in the next section.


\section{Consequences of Equivalence Between Learning Algorithms}
\label{transfer}

We now  study the consequences of the algorithmic equivalences defined in the previous section. In particular we investigate wether these notions of equivalence allow the transfer of learning properties from one learning algorithm to another. We first begin by the following proposition which presents a main of the weak equivalence.

\begin{proposition}\label{prop weak eq}
Let $\mathbf{A}(\lambda)$ and $\mathbf{B}(\lambda)$ two weakly equivalent regularized learning algorithms. Then
$$\forall Z \subset \x \times \y,\quad \inf_{\lambda \in \R} \mathbf{A}(\lambda)(Z)=\inf_{\lambda \in \R} \mathbf{B}(\lambda)(Z).$$
\end{proposition}

\bpf This Proposition directly follows from Definition \ref{weak equivalence}.

Proposition~\ref{prop weak eq} means that the optimal solutions given by two weakly equivalent (regularized) learning algorithms are the same. However, without further assumptions, weak equivalence is of little consequence to the transfer of learning properties from one to the other, such as stability, consistency or generalization bounds. Indeed, these properties are defined for a varying training set either by altering it (such as in stability) or by making it increasingly large (such as in consistency). To illustrate this idea, we will address in particular the question whether weakly equivalence allows or not the transfer of stability.


\textbf{Transfer of stability.} In the following we choose to focus on uniform stability which is an important  property of a learning rule that allows to get bounds on the generalization performance of learning algorithms. Following \cite{Bousquet}, the  uniform stability of a regularized learning algorithm is defined as follows.

\textbf{Notation.} $\l:\y \times \y \mapsto \R_+$ denotes a loss function on $\y$, and $\forall Z \in \z$, $\forall 1 \le i \le \vert Z \vert$, $Z^i$ denotes the set $Z$ minus its $i$-th element.

\begin{definition}\label{def stab}
(Uniform stability). Let $\beta : \N^+ \times \R^*_+ \mapsto \R_+$ be such that $\forall \lambda >0, \lim_{n \rightarrow \infty} \beta(n,\lambda)=0$. A regularized learning algorithms $\a$ is said to be $\beta$- uniformly stable with respect to $\l$ if $$\forall \lambda\! \in\! \R^*_+, \quad \forall Z\! \in \! \z \quad \forall 1 \le \! i \! \le \! n \quad \forall (x,y) \in \x \times \y \quad \vert \l(y,\a(\lambda,Z)) - \l(y,\a(\lambda,Z^i)) \vert \le \beta(\vert Z \vert, \lambda).$$

\end{definition}

We now give an example to show that  weak equivalence is not a sufficient condition for the transfer of uniform stability.

  \textbf{Example:} Let $\a(\cdot)$ denotes the KRR as defined by~\eqref{KRR}  and $\b(\cdot)$ denotes a modified KRR where the regularization term is $\lambda / \vert Z \vert$ instead of $\lambda$. $\a$ and $\b$ are weakly equivalent with $\Phi_{\a \rightarrow \b}(\lambda,Z)=\lambda  \vert Z \vert$.
Under some widely used hypotheses on the kernel and on the output random variable $Y$~\cite{Bousquet}, $\A$ is known to be $\beta$ uniformly stable with $ \beta(n,\lambda)={C_1(1+C_2/\sqrt{\lambda})}/{n \lambda}$ where $C_1$ and $C_2$ are constants, and $n$ is the size of the training set $Z$ (see e.g.  \cite{Bousquet} or \cite{audiffren2013} for more details). Similarly, it is easy to see that $\b$ satisfies the same property but with $\displaystyle \beta(n, \lambda)= {C_1(1+C_2/ \sqrt{ \lambda n })}/{ \lambda}$, which no longer tends to $\infty$ as $n$ increases. Indeed, the regularization term in the learning algorithm $\b(\lambda)$ decreases as  $\vert Z \vert$ increases, and this leads to a decrease of the stability of the learning algorithm.

 It is important to note that if $\a$ and $\b$ are strongly equivalent, then unlike in the weak equivalence case, many properties of $\b$ are  transferred to $\a$. The following Lemma illustrates this idea in the case of stability. 

\begin{lemma}
If $\a$ and $\b$ are strongly equivalent and if $\b$ is $\beta(\cdot,\cdot)$ uniformly stable, then $\a$ is $\beta(\cdot,\Phi_{\a \rightarrow \b}(\cdot))$ uniformly stable.
\end{lemma}

\bpf
Let $(x,y) \in \x \times \y$, $Z \in \z$ and $\lambda \in \R^*_+$. It is easy to see that
\begin{align*}
\vert \l\!\left(y,\a(\lambda,\! Z)(x)  \right)& \!-\!\l\!\left(y,\a(\lambda,\! Z^i)(x)\right)\! \vert \! =\! \vert \l\!\left(y,\b(\Phi_{\a \rightarrow \b}(\lambda),Z)(x)  \right)\! -\!\l\!\left(y,\b(\Phi_{\a \rightarrow \b}(\lambda),Z^i)(x)\right)\! \vert,\\
& \le \beta(\vert Z \vert, \Phi_{\a \rightarrow \b}(\lambda)).\quad\quad\quad\quad\quad\quad\quad\quad\quad\quad\quad\quad\quad\quad\quad\quad\quad\quad\quad\quad \text{ \epf}
\end{align*}

We have shown that, contrarily to strong equivalence, weak equivalence is not sufficient to ensure the transfer of learning properties such as uniform stability.

 In the following, we introduce two additional assumptions which are  a sufficient condition for the transfer of the uniform stability under the weak equivalence. In order to clearly express these assumptions, we first introduce a metric on training set, that is to say a metric \textit{on unordered sequences of different lengths}. To the best of our knowledge, this is a new metric, which allows to easily express learning properties such as stability. We will refer to this metric as the generalized Hamming metric\footnote{ Note that the Hamming metric and the generalized Hamming metric do not coincide on the set of ordered sequences.} (see e.g. \cite{bookash} for more details on the usual Hamming metric).  

\begin{definition}\label{def hamming}
Let $n>0$, $Z^1=\left\lbrace z^1_1, \ldots, z^1_n \right\rbrace$ and $Z^2=\left\lbrace z^2_1, \ldots, z^2_n \right\rbrace$. Let $\Sigma(n)$ denotes the set of all the permutations of  $\left\lbrace 1, \ldots, n \right\rbrace$ and $\underline{H}$ denotes the usual Hamming metric on sequences. $\forall \sigma \in \Sigma(n)$, we denote by $Z^1_\sigma$ the sequence of $n$ elements, whose $i$-th element is $z^1_\sigma(i)$. We define
\begin{itemize}
\item $G_n : (\x\times \y) ^n  \times (\x\times \y) ^n  \mapsto \R^+,$ such that $G_n(Z^1,Z^2)=\min_{\sigma \in \Sigma(n)} \underline{H}(Z^1_\sigma, Z^2)$,
\item  $\H : \z \times \z \mapsto \R^+,$ such that
\begin{equation}
\H(Z_1,Z_2)=\left\lbrace
\begin{aligned}
&\#(Z_1) -\#(Z_2) +\!\!\! \min_{Z \subset Z_1, \#(Z)=\#(Z_2)}\!\! G_{\#(Z_2)}(Z,Z_2) \quad  \text{if $\#(Z_1) \ge \#(Z_2)$},  \\  
&\#(Z_2) -\#(Z_1) +\!\!\! \min_{Z \subset Z_2, \#(Z)=\#(Z_1)}\!\! G_{\#(Z_1)}(Z,Z_1) \quad  \text{ otherwise.}\\
\end{aligned}
\right.
\end{equation}
\end{itemize} 
\end{definition}

The idea of this metric is to consider the number of deletion (i.e. removing an element), insertion~(adding an element) and change~(changing the value of one element)  that allows to move from one training set to  another~(permutations of two elements among a training set are free). The following proposition proves that $\H$ is indeed a metric on $\z$.

\begin{proposition}\label{prop:hamming}
The function $\H : \z \times \z \mapsto \R$ defined in definition \ref{def hamming} is a metric over $\z$.
\end{proposition}

\bpf see the supplementary material.  \epf

\vspace{-0.5cm}


\begin{remark}
The generalized Hamming metric can be used to reformulate the notion of stability. For instance, $\A$ is $\beta$ uniformly stable if and only if $\A$ is $\beta$ lipschitz with respect to the metric $\H$.
\end{remark}

With the help of the metric $\H$, we now introduce two assumptions on the regularity of the functions $\Phi_{\a \rightarrow \b}$ and $\a$. 

\begin{hypothesis}\label{hyp:lips}
$\Phi_{\a \rightarrow \b}(\lambda,\cdot)$ is $C$ Lipschitz decreasing with respect to $\H$, i.e. $\exists c>0$ and $C : \N \mapsto \R$, decreasing, $\lim_{n \rightarrow \infty}C(n)=0$, such that 
\begin{enumerate}
\item $\forall \lambda \in \R^*_+, \quad\forall Z_1,Z_2 \in \z, \quad \vert \Phi_{\a \rightarrow \b}(\lambda,Z_1) - \Phi_{\a \rightarrow \b}(\lambda,Z_2) \vert \le  c \H(Z_1,Z_2)$,
\item $\forall \lambda \in \R^*_+, \quad\forall \in \z, \quad \forall 1 \le i \le \vert Z \vert,  \quad \vert \Phi_{\a \rightarrow \b}(\lambda,Z) - \Phi_{\a \rightarrow \b}(\lambda,Z^i) \vert \le  C(\vert Z \vert)$.
\end{enumerate}

\end{hypothesis}

\begin{hypothesis}\label{hyp:lips2}
Let $\gamma>0$. $\a$ is $\gamma$ Lipschitz with respect to its first variable, i.e.
$\forall Z \in \z,$ $\forall \lambda_1, \lambda_2 \in \R^*_+,$ $\| \a(\lambda_1)(Z) - \a(\lambda_2)(Z) \|_\h \le \gamma \vert \lambda_1 - \lambda_2\vert$.
\end{hypothesis}

These two assumptions are a sufficient condition to the transfer of stability in the weak equivalence case, as shown in the following Proposition.

\begin{proposition}\label{prop weak contraction}
Let $\a$ and $\b$ be two weakly equivalent regularized learning algorithms satisfying Assumptions \ref{hyp:lips} and \ref{hyp:lips2}. Moreover, let $\beta$ be as in Definition \ref{def stab} and locally Lipschitz with respect to its second variable. Suppose that $\exists \kappa>0$ such that $\forall x \in \x, \forall f \in \h$, $\| f(x) \|_\y \le \kappa \|f \|_\h.$   Then:
\begin{center}
 If $\b$ is $\beta$ uniformly stable, then $\a$ is $\beta'$ uniformly stable with $\forall \lambda\! \in\! \R^*_+,$ $\beta'(\cdot,\lambda) = O(\beta(\cdot,\lambda) + C(\cdot) )$.
\end{center}
\end{proposition}

\bpf
Let $(x,y) \in \x \times \y$, $Z \in \z$, $n=\vert Z \vert$ and $\lambda \in \R^*_+$. First note that since $\l$ is $\sigma$-admissible, by using $\lambda'=\Phi_{\a \rightarrow \b}(\lambda,Z)$ and $\lambda''=\Phi_{\a \rightarrow \b}(\lambda,Z^i),$
\begin{align*}
\vert \l&\left(y,\a(\lambda,Z)(x)  \right) -\l\left(y,\a(\lambda,Z^i)(x)\right) \vert = \vert \l\left(y,\b(\lambda',Z)(x)  \right) -\l\left(y,\b(\lambda'',Z^i)(x)\right) \vert,\\
& \le \vert \l\left(y,\b(\lambda',Z)(x)  \right) -\l\left(y,\b(\lambda',Z^i)(x)\right) \vert + \vert \l\left(y,\b(\lambda',Z^i)(x)  \right) -\l\left(y,\b(\lambda'',Z^i)(x)\right) \vert,\\
 & \le \beta(n,\lambda') + \sigma \kappa \| \b(\lambda',Z^i)- \b(\lambda'',Z^i)\|_\h, \le \beta(n) + \sigma \kappa \gamma \vert \lambda'-\lambda'' \vert,\\
& \le \beta(n,\lambda)+\delta 
C(n)  + \sigma \kappa \gamma C(n),
\end{align*}
where in the last line we used the fact that $\beta$ is locally Lipschitz with respect to $\lambda$, hence $\exists \delta(\lambda,\vert \lambda -\lambda' \vert)>0$ such that $\beta(n,\lambda') \le \beta(n,\lambda) + \delta \vert \lambda - \lambda' \vert $. Now, since $\Phi_{\a \rightarrow \b}$ is $C$-Lipschitz decreasing,  $\vert \lambda' - \lambda \vert \rightarrow_{n \rightarrow \infty} 0$, hence the conclusion.
\epf

\vspace{-0.5cm}

\begin{remark} Proposition~\ref{prop weak contraction} can be extended to some non $\sigma$-admissible loss such as the the square loss by using the same ideas as in \cite{Bousquet} and \cite{audiffren2013}.
\end{remark}

The next section is devoted to present an in-depth case study of weak equivalence. It introduces a new regularized learning algorithm, M-RLSR,  and studies the equivalence between KRR and \mbox{M-RLSR}.

 
\section{Case Study: M-RLSR}
\label{MRLSR}

\vspace{0.3cm}

\textbf{Notation. }
In this section, $m > 0$ is a real number, $\x$ a Hilbert space, $\y =\R$, $\h\subset \R^\x$ a separable reproducing kernel Hilbert space~(RKHS),  and $k : \x \times \x \rightarrow \R$ its  positive definite kernel.
For all set of $n$ elements of $\x \times \R$, we denote by $Z=\left\{ (x_1,y_1),..,(x_n,y_n)\right\}$ the training set, and by $K$ the Gram matrix associated to $k$ for $Z$  with~$(K_Z)_{i,j}= k(x_i,x_j)$. Finally, let $Y=(y_1,...,y_n)^\top$ be the output vector. 

The algorithm we investigate here combines a least squares regression with an RKHS regularization term raised to the power of $m$. Formally, we would like to solve the following optimization problem:
\begin{equation}\label{eq Start}
\begin{aligned}
f_Z=\argmin_{f \in \h} \frac{1}{n}\sum_{i=1}^n (y_i - f(x_i))^2 +  \lambda \|f\|^m_\h,
\end{aligned}
\end{equation}
where $m$ is a suitable chosen exponent.
Note that  the classical kernel ridge regression~(KRR) algorithm~\cite{saunders98} is recovered for $m=2$.  This problem has been studied from a theoretical point of view~(see \cite{Mendelson10,Steinwart09}), and in this section we propose a practical way to solve it. 
The problem \eqref{eq Start} is well posed for $m>1$. 
We now introduce a novel $m$-power RLS regression algorithm, generalizing the kernel ridge regression algorithm to an arbitrary regularization exponent.


\subsection{M-RLSR Algorithm}\label{sect theory}

It is worth recalling that the minimization problem \eqref{eq Start} with $m=2$ becomes a standard kernel ridge regression, which has an explicit analytic solution. 
In the same spirit, the main idea of our algorithm is to derive analytically from  \eqref{eq Start} 
a reduced one-dimensional problem on which we apply a root-finding algorithm.

By applying the generalized Representer Theorem from \cite{Dinuzzo12}, we obtain that the solution of \eqref{eq Start} can be written as $
 f_Z= \sum_{i=1}^n \alpha_i k(.,x_i),$
with $\alpha_i \in \R$. The following theorem gives an efficient way to compute the vector $\alpha=(\alpha_1,...,\alpha_n)^\top.$

  \begin{table*}[t]
  \begin{center}
  \begin{tabular}{l}
    \hline \\[-1.0em]
    \textbf{Algorithm 1} \  \  \ $M$-Power RLS Regression Algorithm (M-RLSR) \hspace{3.9cm} \\
    \hline\\[-0.5em]
    \textbf{Input:}  training data $Z=\left\{ (x_1,y_1),..,(x_n,y_n)\right\}$, parameter $\lambda \in \mathbb{R^*_+}$, exponent $m \in \mathbb{R^*_+}$ \\[0.1cm]
    $\quad$ 1.\  \  \textbf{Kernel matrix:}  Compute the Gram matrix, $ K = \big(k(x_i,x_j)\big)_{1\leq i,j\leq n}$  \\[0.2cm]
    $\quad$ 2.\  \  \textbf{Matrix diagonalization:}  Diagonalize $K$ in an orthonormal basis \\[0.1cm]
   \hspace{4cm} $K=Q D Q^\top \quad ; \quad d_i = D_{ii} \ , \ \forall 1\le i\le n $  \\[0.2cm]
    $\quad$ 3.\  \  \textbf{Change of basis:} Perform a basis transformation \\[0.1cm]
   \hspace{4cm} $Y = Q^\top Y  \quad ; \quad y_i = Y_{i} \ , \ \forall 1\le i\le n $  \\[0.2cm]
    $\quad$ 4.\  \  \textbf{Root-finding:} Find the root $C_0$  of the function $F$ defined in \eqref{eq definition F}  \\[0.2cm]
   
   $\quad$ 5.\  \  \textbf{Solution:}  Compute $\alpha$ from \eqref{eq def alpha} and reconstruct the weights \\[0.1cm]
   \hspace{4cm}   $\displaystyle (\alpha_i)_{1\leq i \leq n}= \frac{2y_i}{2d_i +  \lambda m n C_0} \ \ $ and $\ \ \alpha = Q \alpha$  \\[0.3cm]
    \hline
  \end{tabular}
    \end{center}
  \end{table*}

\begin{theorem}\label{th algo}
Let $Q$ an orthonormal matrix and $D$ a diagonal matrix such that $K=QDQ^\top$. Let $y_i'$ be the coordinates of $Q^ T Y$, $(d_i)_{1\le i \le n}$ the elements of the diagonal of $D$,  $C_0 \in \R_+$ and $m>1$.
Then the vector $\alpha= Q \alpha'$ with 
\begin{equation}
\label{eq def alpha}
\alpha'_i= \frac{2y_i'}{2d_i + \lambda m n C_0  } \ , \  \forall 1 \le i \le n \ , \quad 
\end{equation}
is the solution of \eqref{eq Start} if and only if
$C_0$ is the root of the function $F:\R_+ \rightarrow \R$ defined by
\begin{equation}\label{eq definition F}
\begin{aligned}
F(C)&=\big(\sum_{i=1}^n   \frac{4d_i  y_i'^2}{(2d_i + \lambda m n C)^2  }\big )^{m/2 - 1} - C.\\
\end{aligned}
\end{equation}
\end{theorem}

\bpf
The proof of Theorem \ref{th algo} can be found in the supplementary material. \epf

It is important to note that for $m>1$, $F$ has a unique root $C_0$ and that since $F$  is a function from $\R$ to $\R$, computing $C_0$ using a root-finding algorithm, e.g. Newton's method, is a fast and accurate procedure.
Our algorithm uses these results to provide an efficient solution to regularized least squares regression with a variable regularization exponent~$m$~(see Algorithm 1).

%

\subsection{Equivalence Between M-RLSR and KRR}

Here we will show that M-RLSR and KRR are only weakly equivalent but not strongly equivalent. The idea is that, when $m>1$, the objective function of the \mbox{M-RLSR} optimization problem~\eqref{eq Start} is strictly convex, and then by Lagrangian duality it is equivalent to its unconstrained version. The weak equivalence is proved in the following proposition.

\begin{proposition}\label{lemma Z equivalence}
$\forall m>1,$ $\forall Z \in \z,\exists F_{Z,m}: \R^+ \to \R^+$, bijective, such that  $\forall \lambda >0$,  M-RLSR with regularization parameter $\lambda$ and KRR with regularization parameter $ \lambda_2 = F_{Z,m}(\lambda)$ are weakly equivalent. Moreover,
$$ \Phi_{\a \rightarrow \b}(\lambda,Z)= \frac{m \lambda}{2} C_0(Z,m,\lambda),$$
where $C_0(Z,m,\lambda)$ is the unique root of the function $F$ defined in \eqref{eq definition F}.
\end{proposition}

\bpf 
For $m>1$, the equivalence between constrained and unconstrained strictly convex optimization problems~\cite[Appendix~A]{Kloft11} implies that  
$\exists \Gamma_{m,Z,\lambda} >0$ such that the minimization problem defined by \eqref{eq Start} on $Z$ it is equivalent to the following constrained problem: 
\begin{equation}
\nonumber
\begin{aligned}
&\argmin_{f \in \h} \frac{1}{\vert Z \vert}\sum_{(x,y) \in Z} (y - f(x))^2,
&\text{s.t. } \| f \|_\h ^m \le \g.
\end{aligned}
\end{equation}
The constrain is equivalent to $\| f \|_\h ^2 \le \g^{2/m}$, thus we deduce that $\exists \lambda_2(m,Z,\lambda) >0$ such that \eqref{eq Start} with regularization parameter $\lambda$ is equivalent to
\begin{equation}
\nonumber
\begin{aligned}
&\argmin_{f \in \h} \frac{1}{\vert Z \vert}\sum_{(x,y) \in Z} (y - f(x))^2 + \lambda_2(m,Z,\lambda) \|f \|^2_\h,
\end{aligned}
\end{equation}
i.e., the KRR minimization problem with a regularization parameter $\lambda_2(m,Z,\lambda)$. Hence M-RLSR with $\lambda$ is  weakly equivalent to KRR with $\lambda_2(m,Z,\lambda)$. It is easy to see from~\eqref{eq def alpha} that the function $F_{Z,m}$ that maps $\lambda$ to the corresponding $\lambda_2$ has the form $ F_{Z,m}(\lambda):= \frac{m}{2} C_0(Z,m,\lambda) \lambda$.\epf

Since $\Phi_{\a \rightarrow \b}(\lambda,Z)$ heavily depends on Z,  M-RLSR and  KRR are not strongly equivalent but only weakly equivalent. 
%
Moreover, Assumptions 1 and 2 are not satisfied in this case, hence stability of M-RLSR cannot be deduced from that of KRR. A stability analysis of M-RLSR can be found in the supplementary material.

 \begin{figure}
 \centering
 \includegraphics[scale=0.27]{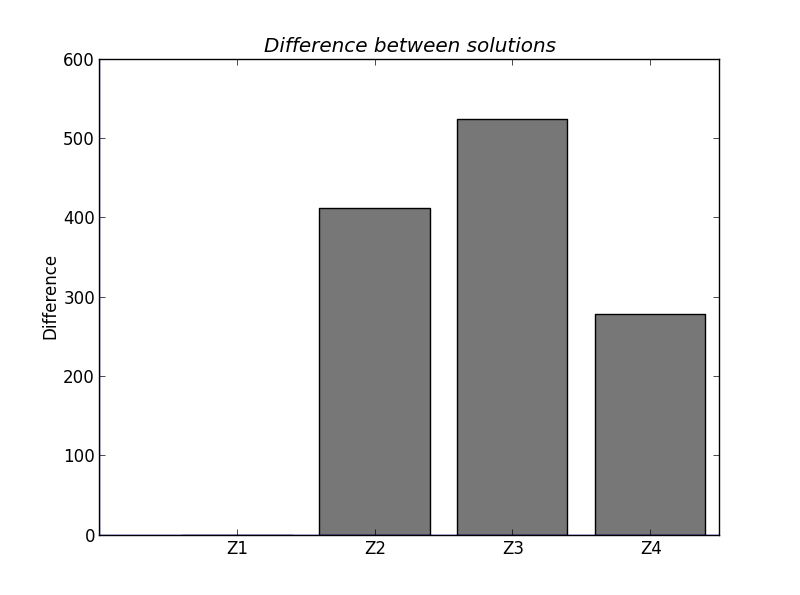}
 $\quad$
 \includegraphics[scale=0.27]{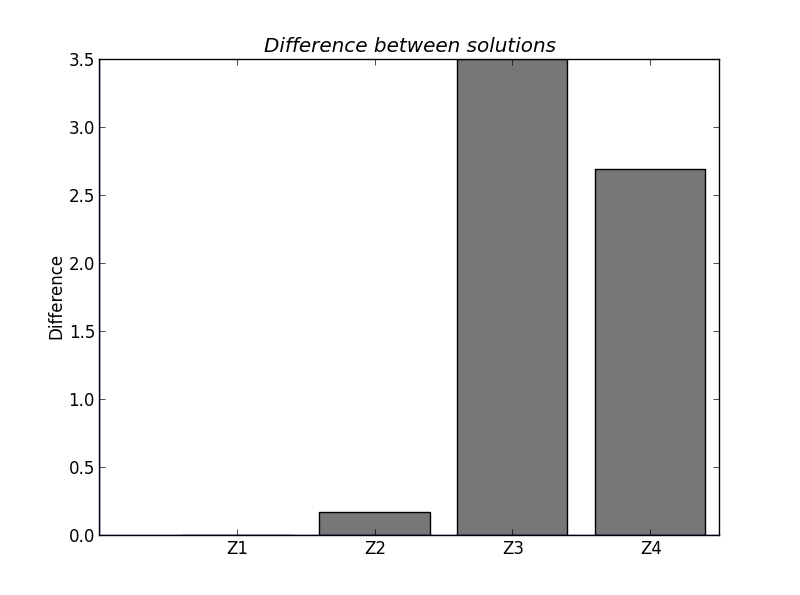}
 \caption{ The norm of the difference between the optimal solutions given by M-RLSR and KRR on two datasets randomly split into 4 parts $Z_1, \ldots, Z_4$.  While the difference on $Z_1$ is zero for the two algorithms since they are weakly-equivalent, they give different solution for $Z_1$, $Z_2$ and $Z_3$. (left) Concrete compressive strength dataset: $m=1.5,\lambda=1e-2$ (obtained by 10-fold cross validation) and $\lambda_2=5.6e-4$ (computed
from Proposition \ref{lemma Z equivalence}). (right) Synthetic dataset: $m=1.2,\lambda=1e-5$ (obtained by 10-fold cross validation) and $\lambda_2=6.5e-7$ (computed from the Proposition \ref{lemma Z equivalence}).}
 \label{fig nonequiv}
 \end{figure}

\subsection{Experiments on Weak Equivalence}\label{sect Xp}

In this subsection, we conduct experiments on synthetic and real-world datasets to illustrate the fact that \mbox{M-RLSR} and KRR algorithms are only weakly equivalent but not strongly equivalent. 
We use the Concrete Compressive Strength (1030 instances, 9 attributes) real-world dataset extracted from the UCI repository\footnote{ \url{http://archive.ics.uci.edu/ml/datasets}.}. Additionally, we also use a synthetic dataset (2000 instances, 10 attributes) described in \cite{tsang05}. In this dataset, inputs~$(x_1, . . . , x_{10})$ are generated independently and uniformly over $\left[ 0, 1\right]$ and outputs are computed from $y = 10 \sin(\pi x_1 x_2)+ 20(x_3 - 0.5)^2 + 10x_4 + 5x_5 + \mathcal{N}(0, 1).$ 

We randomly split these datasets into 4 parts of equal size  $Z_1, \ldots, Z_4$. Using $Z_1$, $m$ is fixed and the regularization parameter $\lambda$ is chosen by a 10-fold cross-validation for \mbox{M-RLSR}. Then the equivalent $\lambda_2$ for KRR is computed using Proposition~\ref{lemma Z equivalence}. For each part $Z_i, 1\leq i \leq 4$, we calculate the norm of the difference between the optimal solutions given by M-RLSR and KRR.
The results are presented in Figure~\ref{fig nonequiv}. The difference between the solutions of the two algorithms is equal to $0$ on $Z_1$, but since both algorithms are only weakly equivalent, the difference is strictly positive on $Z_2,Z_3,Z_4$, showing that the algorithms are not strongly equivalent.
Additional experiments regarding the M-RLSR and its algorithmic properties can be found in the supplementary material.


\section{Conclusion}\label{sect End}

We have presented a novel way of theoretically analyzing and interpreting relations between machine learning algorithms, namely the concept of algorithmic equivalence. More precisely, we have proposed two notions of equivalence of learning algorithms, weak and strong equivalence, and we have shown how to use them to transfer learning properties, such as stability, from one learning algorithm to another.

Although this work has focused in particular on the transfer of stability using the concept of algorithmic equivalence, we believe that it can be extended to study the transfer of other algorithmic properties such as sparsity, robustness and generalization. Future work will also aim at further quantifying the equivalence relations introduced by providing efficient tools that can help  to decide whether two learning algorithms are weakly or strongly equivalent.

\newpage
\bibliographystyle{unsrt}
\bibliography{biblioinfo}

\newpage
\appendix

\section{Weak Equivalence and Regularization Path}

Let $Z \in \z$ be a fixed training set, and $\mathbf{A}$ and $\mathbf{B}$ two weakly equivalent algorithm. By definition of the weak equivalence $\mathbf{A}(\cdot)(Z)= \mathbf{B}(\Phi_{\a \rightarrow \b}(\cdot,Z))(Z)$. This formulation highlights the consequence of the weak equivalence with the regularization path: the regularization path of $\mathbf{B}$ can be obtained from the the regularization path of $\mathbf{A}$  with the bijective transformation $\Phi_{\a \rightarrow \b}(\cdot,Z)$ of the variable $\lambda$.  It is important to note that $\Phi_{\a \rightarrow \b}(\cdot,Z)$ depends on $Z$, i.e. \textit{the relation between the regularization path of $\mathbf{A}$ and $\mathbf{B}$ depends on $Z$}.
The same can be said for the error curves, but  Proposition~4.1 ensure that they share the same minimum value (see figure 1).

\section{Proof of Proposition 4.6}

\bpf It is easy to see that $\H$ is symmetric and $\H(Z_1,Z_2) \ge 0$ and is equal to $0$ if and only if $Z_1=Z_2$. Now, in order to prove the sub-additivity of $\H$, let $Z_1, Z_2$ and $Z_3 \in \z$. 
Note that $G_n$ counts the number of elements which differs between two unordered sequences, and thus is sub-additive.

We only write here the case $\#(Z_2) \ge \#(Z_1) \ge  \#(Z_3)$, the other cases are done likewise. Let for $i=1,3$ $$\displaystyle Z^2_i= \argmin_{Z \subset Z_2, \#(Z)=\#(Z_i)} G_{\#(Z_i)}(Z,Z_i).$$
Without any loss of generality, suppose that $\#(Z^2_1) \ge \#(Z^2_3)$. Then, 
\begin{equation}\label{eq:ineg1}
 \#(Z_2) \ge \#(Z^2_1) + G(\hat{Z},Z^2_3)\text{, where }
\hat{Z}= \argmin_{Z \subset Z^2_1, \#(Z)=\#(Z^2_3)} G_{\#(Z^2_3)}(Z,Z^2_3).
\end{equation}

\begin{align*}
\H(Z_1,Z_2)\!+\!\H(Z_2,Z_3) &=\! \#(Z_2)\! -\!\#(Z_1)\! +\!  G_{\#(Z_1)}(Z^2_1,Z_1)\! +\!\#(Z_2)\! -\!\#(Z_3) + G_{\#(Z_3)}(Z^2_3,Z_3) \\
&\ge\! \#(Z_2)\! -\!\#(Z_3)\! +\! G_{\#(Z_1)}(Z^2_1,Z_1)\! +\! G_{\#(Z_3)}(Z^2_3,Z_3)+ G_{\#(Z_3)}(\hat{Z},Z^2_3)\\
&\ge \#(Z_2) -\#(Z_3) + G_{\#(Z_1)}(Z^2_1,Z_1) + G_{\#(Z_3)}(\hat{Z},Z_3)\\
&\ge \#(Z_2) -\#(Z_3) + \min_{Z \subset Z_1, \#(Z)=\#(Z_3)} G_{\#(Z_3)}(Z,Z_3)\\
&\ge \#(Z_1) -\#(Z_3) + \min_{Z \subset Z_1, \#(Z)=\#(Z_3)} G_{\#(Z_3)}(Z,Z_3)= \H(Z_1,Z_3)\\
\end{align*}

where, in the second line we used \eqref{eq:ineg1}, third line we used the sub-additivity of $G$, fourth line we used the fact that $\hat{Z} \subset Z^2_1$, and last line we used $\#(Z_2) \ge \#(Z_1) \ge  \#(Z_3).$ \epf

\section{Proof of Theorem 1}

Remember that we are trying to solve the following problem :
\begin{equation}\label{eq Start}
\begin{aligned}
f_Z=\argmin_{f \in \h} \frac{1}{n}\sum_{i=1}^n (y_i - f(x_i))^2 +  \lambda \|f\|^m_\h,
\end{aligned}
\end{equation}
where $m$ is a suitable chosen exponent.

For the convenience of the reader, let us rewrite the theorem we are going to prove
 \begin{figure}
 \centering
\includegraphics[scale=0.3]{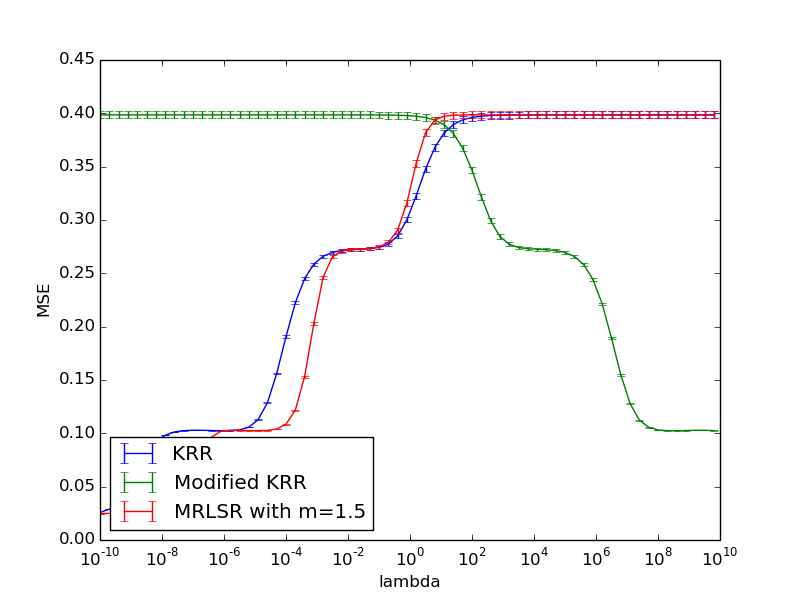}
\includegraphics[scale=0.3]{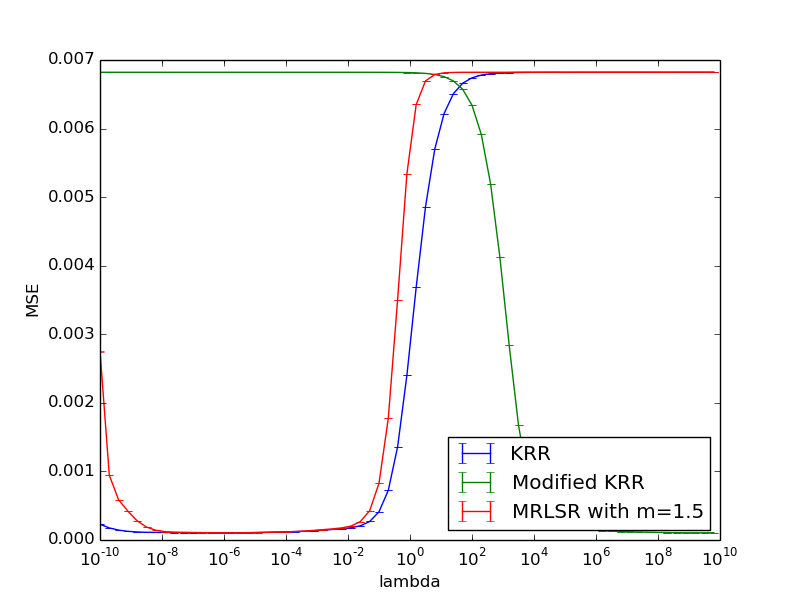}
 \caption{ Error curves for  KRR, modified KRR defined by \eqref{KRR-hand} and  M-RLSR with m=1.5 on the dataset Yatch hydrodynamic(left) and (right), both extracted from the UCI repository, }
 \label{fig error curve}
 \end{figure}
\begin{theorem}\label{th algo}
Let $Q$ an orthonormal matrix and $D$ a diagonal matrix such that $K=QDQ^\top$. Let $y_i'$ be the coordinates of $Q^ T Y$, $(d_i)_{1\le i \le n}$ the elements of the diagonal of $D$,  $C_0 \in \R_+$ and $m>1$ .
Then the vector $\alpha= Q \alpha'$ with 
\begin{equation}
\label{eq def alpha}
\alpha'_i= \frac{2y_i'}{2d_i + \lambda m n C_0  } \ , \  \forall 1 \le i \le n \ , \quad 
\end{equation}
is the solution of \eqref{eq forme f_Z} if and only if
$C_0$ is the root of the function $F:\R_+ \rightarrow \R$ defined by
\begin{equation}\label{eq definition F}
\begin{aligned}
F(C)&=\big(\sum_{i=1}^n   \frac{4d_i  y_i'^2}{(2d_i + \lambda m n C)^2  }\big )^{m/2 - 1} - C.\\
\end{aligned}
\end{equation}
\end{theorem}

\bpf

First notice that, the objective function to minimize is G\^{a}teaux differentiable in every direction. Thus, since $f_Z$ is a minimum, we have:
\begin{equation}\nonumber
\begin{aligned}
0= \sum_{i=1}^n -2 k(.,x_i) ( y_i - f_Z(x_i))  +  \lambda m n \|f_Z\|^{m-2}_\h f_Z,
\end{aligned}
\end{equation}
i.e.,
\begin{equation}\nonumber
\begin{aligned}
 f_Z= \sum_{i=1}^n 2 k(.,x_i) \frac{ y_i - f_Z(x_i)}{\lambda m n \|f_Z\|^{m-2}_\h}.
 \end{aligned}
\end{equation}
That is to say, $f_Z$ can be written in the following form: 
\begin{equation}\label{eq forme f_Z}
\begin{aligned}
 f_Z= \sum_{i=1}^n \alpha_i k(.,x_i),
 \end{aligned}
\end{equation}
with $\alpha_i \in \R$.
Notice, that we have recovered exactly the form of the representer theorem, which can also be derived from a result due to Dinuzzo and Sch\"olkop~\cite{Dinuzzo12}. Now by combining~\eqref{eq Start} and~\eqref{eq forme f_Z}, the initial problem becomes
\begin{equation}\label{equation depart dual}
\begin{aligned}
\alpha=\argmin_{a \in \R^n} ( Y - K a ) ^\top ( Y - K a ) + n \lambda (a^\top K a )^{m/2},
\end{aligned}
\end{equation}
where $\alpha =(\alpha_i)_{1 \le i \le n}$ is the vector to determine. The following theorem gives an explicit formula for $\alpha$ that solves the optimization problem  \eqref{equation depart dual}.

By computing the G\^{a}teaux derivative of the objective function to minimize in \eqref{equation depart dual},  we obtain that $\alpha$ must verify 
\begin{equation}\label{eq before K inverse}
\nonumber
\begin{aligned}
Y&=K \alpha +  \lambda  \frac{mn}{2} (\alpha^\top K \alpha )^{m/2 - 1} \alpha.
\end{aligned}
\end{equation}
Then, since $K$ is symmetric and positive semidefinite, $\exists Q$ an orthonormal matrix (the matrix of the eigenvectors) and $D$ a diagonal matrix with eigenvalues $(d_i)_{1 \le i \le n} \ge 0$ such that $K = Q D Q^\top$. Hence,
\begin{equation}\nonumber
\begin{aligned}
Y&=Q D Q^\top \alpha +  \lambda  \frac{mn}{2} ((Q^\top \alpha)^\top D  (Q^\top \alpha) )^{m/2 - 1} \alpha\\
\Rightarrow Q^\top Y&=D Q^\top \alpha +  \lambda  \frac{mn}{2} ((Q^\top \alpha)^\top D  (Q^\top \alpha) )^{m/2 - 1} Q^\top \alpha.
\end{aligned}
\end{equation}
Given this, one can define a new representation by changing the basis such that $Y'=Q^\top Y$ and $\alpha'= Q^\top \alpha$. We obtain
\begin{equation}\nonumber
\begin{aligned}
Y'&=D \alpha' +  \lambda  \frac{mn}{2} (\alpha'^\top D   \alpha' )^{m/2 - 1}  \alpha'.\\
\end{aligned}
\end{equation}
Now if we write the previous equation for every coefficient of the vectors, we obtain that 
\begin{equation}\nonumber
\left\{
\begin{aligned}
&y'_i= d_i \alpha'_i +  \lambda \frac{mn}{2} (\sum_{j=1}^n d_j \alpha_j'^2 )^{m/2 - 1}  \alpha_i' \quad , \quad \forall 1 \le i \le n.
\end{aligned}
\right.
\end{equation}
Note that $(\sum_{j=1}^n d_j \alpha_j'^2 )^{m/2 - 1}$ is the same for every equation (i.e. it does not depend on $i$), so we can rewrite the system as follows, where ${C \in \R}$
\begin{equation}\label{Eq valeur alpha fonction C}
\left\{
\begin{aligned}
C&=\big(\sum_{j=1}^n d_j \alpha_j'^2 \big)^{m/2 - 1}\\
&\text{and}\\
&\alpha_i'= \frac{2y'_i}{2d_i +  \lambda mn C  } \quad , \quad  \forall 1 \le i \le n.
\end{aligned}
\right.
\end{equation}

which is well defined if $d_i +  \lambda mn C \neq 0$, which is the case when $C>0$. Since $C \ge 0$ by definition, the only possibly problematic case is $C=0$, but this implies that $Y=0$, which is a degenerated case. Now we just need to calculate $C$, which verifies:
\begin{equation}\label{eq valeur C}
\nonumber
\begin{aligned}
C&=\big(\sum_{i=1}^n d_i \alpha_i'^2\big )^{m/2 - 1}=\big(\sum_{i=1}^n  \frac{4d_i  y_i'^2}{(2d_i +  \lambda mn C)^2  } \big)^{m/2 - 1}.
\end{aligned}
\end{equation}
Thus to obtain an explicit value for $\alpha'$, we need only to find a root of the function $F$ defined as follows~:
\begin{equation}\nonumber
\begin{aligned}
F(C)&=\big(\sum_{i=1}^n  \frac{4d_i  y_i'^2}{(2d_i +  \lambda mn C)^2  } \big)^{m/2 - 1} - C.\\
\end{aligned}
\end{equation}
We have proven that any solution of \eqref{equation depart dual} can be written as a function of $C_0$, a root of $F$.
But for $m>1$, $F$ is strictly concave, and $F(0)>0$, hence it has at most one root in $ \R_+$. Thus since $\lim_{C\rightarrow + \infty} F(C)=-\infty$, $F$ has exactly one root, which proves Theorem~\ref{th algo}.
 \epf
 
 \section{Stability Analysis of M-RLSR}\label{sect Stability}

The notion of algorithmic stability, which is the behavior of a learning algorithm following a change of the training data, was used successfully by Bousquet and Elisseeff~\cite{Bousquet} to derive bounds on the generalization error of  kernel-based learning algorithms.
In this section, we extend the stability results of~\cite{Bousquet} to cover the $m$-power RLSR algorithm. 
We show here that the algorithm is stable for $m\ge2$. 

In this section we denote by \underline{X} and \underline{Y} a pair of random variables following the unknown distribution~$D$ of the data, $\underline{X}$ representing the input and $\underline{Y}$ the output, by $\zi=Z\setminus(x_i,y_i)$ the training set from which was removed the element $i$. Let $c(y,f,x)= (y-f(x))^2$ denotes the cost function used in the algorithm. For all $f\in \h$, let $R_e(f,Z)=1/n \sum_{1\le i\le n} c(y_i,f,x_i)$ be the empirical error and $R_r(f,Z)=R_e(f,Z)+\lambda \|f\|^m_\h$ be the regularized error.
Let us recall the definition of uniform stability.

\begin{definition} 
An algorithm $Z \rightarrow f_Z$ is said $\beta$ uniformly stable if and only if $\forall n \ge 1 $, $\forall 1 \le i \le n,$  $\forall Z $ a realization  of $n$ i.i.d. copies of $(\underline{X},\underline{Y})$,$ \forall (x,y) \in \x \times \y$ a $Z$ independent realization of $(\underline{X},\underline{Y})$,
 we have $\vert c(y,f_Z,x)- c(y,f_\zi,x)\vert \le \beta.$
\end{definition}

To prove the stability of a learning algorithm, it is common to make the following assumptions.

\begin{hypothesis}\label{Hyp Y bounded}
$\exists C_y>0 \text{ such that } \vert\underline{Y}\vert < C_y \text{ a.s.}$
\end{hypothesis}

\begin{hypothesis}\label{Hyp K bounded}
$\exists \kappa>0 \text{ such that }\sup_{x\in \x} k(x,x) < \kappa^2$
\end{hypothesis}

\begin{lemma}\label{lemma c is lips}
If Hypotheses \ref{Hyp Y bounded} and \ref{Hyp K bounded} hold, then $\forall n \ge 1 $, $\forall 1 \le i \le n,$  $\forall Z $ a realization  of $n$ i.i.d. copies of $(\underline{X},\underline{Y})$,$ \forall (x,y) \in \x \times \y$ a $Z$ independent realization of $(\underline{X},\underline{Y})$,
$$\vert c(y,f_Z,x)- c(y,f_\zi,x)\vert \le C \vert f_Z(x) - f_\zi(x) \vert,$$
with $C=2 \left(C_y +  \kappa\left(\frac{C_y ^2}{\lambda}\right)^\frac{1}{m}\right) $.
\end{lemma}

\bpf
Since $\h$ is a vector space, $0 \in \h$, and 

\begin{equation}\nonumber
\begin{aligned}
  \lambda \|f_Z\|^m &\le\frac{1}{n}\sum_{i=1}^n (y_i - f_Z(x_i))^2 +  \lambda \|f_Z\|^m_\h \\
  &\le \frac{1}{n}\sum_{k=1}^n \|y_k-0 \|^2 +\lambda \|0\|^m_\h \le C_y ^2, 
\end{aligned}
\end{equation}
where we used the definition of $f_Z$ as the minimum of \eqref{eq Start} and Hypothesis \ref{Hyp Y bounded}. 
Using the reproducing property and Hypothesis \ref{Hyp K bounded}, we deduce that 
\begin{equation}\label{eq bound f_z(x)}
\nonumber
\vert f_Z(x) \vert \le \sqrt{k(x,x)} \|f_Z\|_\h \le \kappa\|f_Z\|_\h  \le \kappa\left(\frac{C_y ^2}{\lambda}\right)^\frac{1}{m}. 
\end{equation}
The same reasoning holds for $f_\zi$. Finally,
\begin{equation}\nonumber
\begin{aligned}
\vert c(y, & f_Z,x)- c(y,f_\zi,x)\vert \\ &=\vert (y - f_Z(x))^2 -  (y - f_\zi(x))^2 \vert\\
&\le 2 \left(C_y +  \kappa\left(\frac{C_y ^2}{\lambda}\right)^\frac{1}{m}\right)  \vert f_Z(x)-f_\zi(x) \vert .      
\end{aligned}
\end{equation}      \epf

\vspace{-0.3cm}
The stability of our algorithm when $m\ge 2$ is established in the following theorem, whose proof is an extension of Theorem~22 in \cite{Bousquet}. The original proof concerns the KRR case when $m=2$. The beginning of our proof is similar to the original one; but starting from \eqref{eq inequality with t}, the proof is modified to hold for $m \ge 2$, since the equalities used in \cite{Bousquet} no longer holds when $m>2$,. We use inequalities involving generalized Newton binomial theorem instead.

\begin{theorem} \label{beta stable}
Under the assumptions \ref{Hyp Y bounded} and \ref{Hyp K bounded}, algorithm $Z \rightarrow f_Z$ defined in~\eqref{eq Start} is $\beta$ stable $\forall m>=2$ with $$\beta= C \kappa  \left(    2^{m-2}\frac{C \kappa  }{\lambda n}\right)^{\frac{1}{m-1}}.$$
\end{theorem}
\bpf
Since $c$ is convex with respect to $f$, we have $\forall 0 \le t \le 1$
\begin{equation}\nonumber
\begin{aligned}
c(y,f_Z +& t(f_\zi-f_Z),x) -c(y,f_Z,x) \\
 &\le t \left(c(y,f_\zi,x) - c(y,f_Z,x) \right).
\end{aligned}
\end{equation}
Then, by summing over all couples $(x_k,y_k)$ in $\zi$,
\begin{equation}\label{eq Rempun}
\begin{aligned}
R_{e}(f_Z + &t(f_\zi-f_Z),\zi) -R_{e}(f_Z ,\zi) \\
& \le t \left(R_{e}(f_\zi,\zi) -R_{e}(f_Z ,\zi) \right).
\end{aligned}
\end{equation}
By symmetry, \eqref{eq Rempun} holds if $Z$ and $Z_i$ are permuted. By summing this symmetric equation and \eqref{eq Rempun}, we obtain
\begin{equation}\label{eq Remptrois}
\begin{aligned}
&R_{e}(f_Z + t(f_\zi-f_Z),\zi) -R_{e}(f_Z ,\zi)\\
&+ R_{e}(f_\zi + t(f_Z-f_\zi),\zi) -R_{e}(f_\zi ,\zi)  \le 0.
\end{aligned}
\end{equation}
Now, by definition of $f_Z$ and $f_\zi$,
\begin{equation}\label{eq fmin}
\begin{aligned}
&R_{r}(f_Z ,Z) - R_{r}(f_Z + t(f_\zi-f_Z),Z)\\
&+R_{r}(f_\zi ,\zi)-  R_{r}(f_\zi + t(f_Z-f_\zi),\zi)  \le 0.
\end{aligned}
\end{equation}
By using \eqref{eq Remptrois} and \eqref{eq fmin} we get
\begin{equation}\label{eq inequality with t}
\begin{aligned}
&c(y_i,f_Z,x_i) -c(y_i,f_Z + t(f_\zi-f_Z),x_i)\\
&\hspace{1cm} + \lambda n \left( \| f_Z \|^m_\h - \| f_Z + t(f_\zi-f_Z)\|^m_\h\right.\\
&\hspace{1cm}\left.+ \| f_\zi \|^m_\h -\| f_\zi + t(f_Z-f_\zi) \|^m_\h \right)\le 0,
\end{aligned}
\end{equation}
This inequality holds $\forall t \in \left[0,1\right]$. By choosing $t=1/2$ in \eqref{eq inequality with t}, we obtain that
\begin{equation}\label{eq inequality with 1/2}
\begin{aligned}
\vert c(&y_i,f_Z,x_i) -c(y_i,f_Z + \frac{1}{2}(f_\zi-f_Z),x_i) \vert\\
& \ge n \lambda \left( \| f_Z \|^m_\h - 2 \left\| \frac{f_\zi+f_Z}{2}\right\|^m_\h
+ \| f_\zi \|^m_\h  \right),
\end{aligned}
\end{equation}
Let $u=(f_Z+f_\zi)/2$ and $v=(f_Z-f_\zi)/2$. Then,
\begin{equation}\nonumber
\begin{aligned}
&\| u +v \|^m_\h+ \|u-v \|^m_\h - 2 \left\| u\right\|^m_\h
- 2 \left\| v\right\|^m_\h\\
&\hspace{0.1cm}=\| f_Z \|^m_\h+ \| f_\zi \|^m_\h - 2 \left\| \frac{f_\zi+f_Z}{2}\right\|^m_\h
- 2 \left\| \frac{f_\zi-f_Z}{2}\right\|^m_\h\\
&\hspace{0.1cm}=\! \left( \|u\|^2_\h  \! +\! \|v\|^2_\h \! +\! 2 \left\langle u,v \right\rangle_\h \right)^{m/2}-\!2 \left( \| u\|^2_\h\right)^{m/2}\\
&\hspace{0.1cm} +\! \left( \|u\|^2_\h \!+\! \|v\|^2_\h \!-\! 2 \left\langle u,v \right\rangle_\h \right)^{m/2}\!\!
-\!2 \left( \| v\|^2_\h\right)^{m/2}\\
&\hspace{0.1cm}\ge 2\left( \|u\|^2_\h + \|v\|^2_\h \right)^{m/2} -2 \left( \| u\|^2_\h\right)^{m/2}
-2 \left( \| v\|^2_\h\right)^{m/2}\\
&\hspace{0.1cm} \ge 0,
\end{aligned}
\end{equation}
where in the last transition we used both Newton's generalized binomial theorem for the first inequality and the fact that $m/2 >1$ for the second one. Hence, we have shown that
\begin{equation}\label{eq inequality to prove}
\begin{aligned}
\| f_Z \|^m_\h - 2 \left\| \frac{f_\zi+f_Z}{2}\right\|^m_\h
+ \| f_\zi \|^m_\h \ge 2 \left\| \frac{f_\zi-f_Z}{2}\right\|^m_\h.
\end{aligned}
\end{equation}

Now, by combining \eqref{eq inequality with 1/2} and \eqref{eq inequality to prove}, we obtain by using Lemma \ref{lemma c is lips},
\begin{equation}\nonumber
\begin{aligned}
&\|f_Z-f_\zi \|^m_\h\\
 &\hspace{0.1cm}\le \frac{2^{m-1}}{\lambda n} \left( c(y_i,f_Z + \frac{1}{2}(f_\zi-f_Z),x_i)-c(y_i,f_Z,x_i)\right) \\
 &\hspace{0.1cm}\le 2^{m-2} \frac{C  }{\lambda n} \|f_\zi(x_i) - f_Z(x_i)\|_\y\\
  &\hspace{0.1cm}\le  2^{m-2}\frac{C \kappa  }{\lambda n} \|f_\zi- f_Z\|_\h,
\end{aligned}
\end{equation}
which gives that $$\displaystyle \|f_Z-f_\zi \|_\h \le \left(    2^{m-2}\frac{C \kappa  }{\lambda n}\right)^{\frac{1}{m-1}}.$$
This implies that, $\forall (x,y)$ a realization of $(X,Y)$,
\begin{equation}\nonumber
\begin{aligned}
\vert c(y,f_Z,x) - c(y,f_\zi,x) \vert &\le C \|f_Z(x) - f_\zi(x)\|_\y\\
&\le C \kappa  \left(    2^{m-2}\frac{C \kappa  }{\lambda n}\right)^{\frac{1}{m-1}}.
\end{aligned}
\end{equation}\vspace{-0.3cm}\epf

\vspace{-0.4cm}

For $1<m<2$, the problem \eqref{eq Start} is well posed but the question whether the algorithm is stable or not in this case remains open.  Future studies need to be conducted to further address this issue explicitly.

\section{Additional Experiments on M-RLSR}

In this section, we conduct experiments on synthetic and real-world datasets to evaluate 
the efficiency of the proposed algorithm.
%
We use the following real-world datasets extracted from the UCI repository\footnote{ \url{http://archive.ics.uci.edu/ml/datasets}.}: Concrete Compressive Strength (1030 instances, 9 attributes), Concrete Slump Test (103 instances, 10 attributes), Yacht Hydrodynamics~(308 instances, 7 attributes), Wine Quality (4898 instances, 12 attributes), Energy Efficiency (768 instances, 8 attributes), Housing (506 instances, 14 attributes) and Parkinsons Telemonitoring (5875 instances, 26 attributes). Additionally, we also use a synthetic dataset (2000 instances, 10 attributes) described in \cite{tsang05}. In this dataset, inputs~$(x_1, . . . , x_{10})$ are generated independently and uniformly over $\left[ 0, 1\right]$ and outputs are computed from $y = 10 \sin(\pi x_1 x_2)+ 20(x_3 - 0.5)^2 + 10x_4 + 5x_5 + \mathcal{N}(0, 1).$ 
In all our experiments, we use a Gaussian kernel $k_\mu(x,x')=\exp(-\|x-x'\|_2^2 / \mu)$
with $ \mu= \frac{1}{n^2}\sum_{i,j} \|x_i - x_j\|_2^ 2$, and the scaled root mean square error (RMSE), defined by $\frac{1}{\max y_i} \sqrt{\frac{1}{n}\sum_i (y_i - f(x_i))^2 }$, as evaluation measure.

\subsection{Speed of Convergence} \label{subs xp conv}

We compare here the convergence speed of M-RLSR with $m\leq1$ and KRR on  Concrete compressive strength, Yacht Hydrodynamics, Housing, and Synthetic datasets. As before, each dataset is randomly split into two parts (70\% for learning and 30\% for testing). The parameters $m$ and $\lambda$ are selected using cross-validation: we first fix $\lambda$ to 1 and choose $m$ over a grid ranging from 0.1 to 1, then $\lambda$ is set by cross-validation when $m$ is fixed. For KRR, $\lambda_2$ is computed from $\lambda$ and $m$ using  Proposition 5.1.

\begin{figure}
\centering
\includegraphics[scale=0.25]{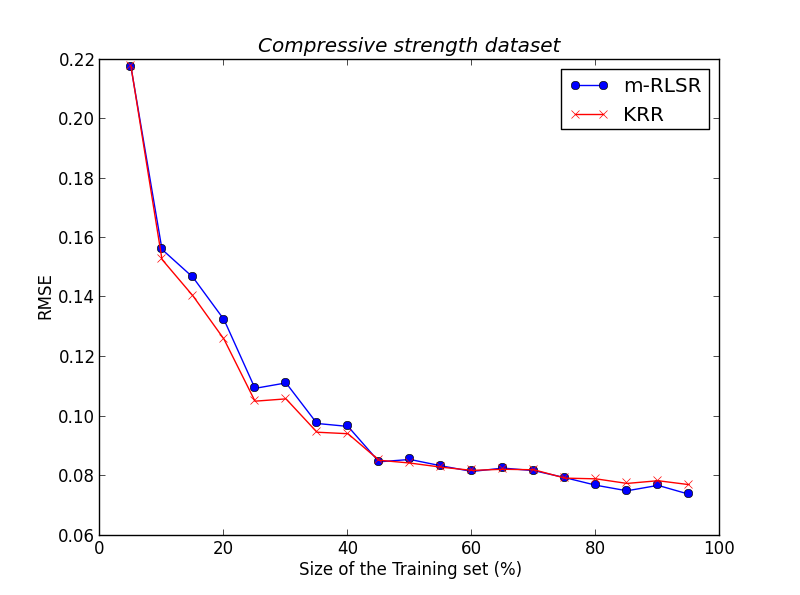}
\includegraphics[scale=0.25]{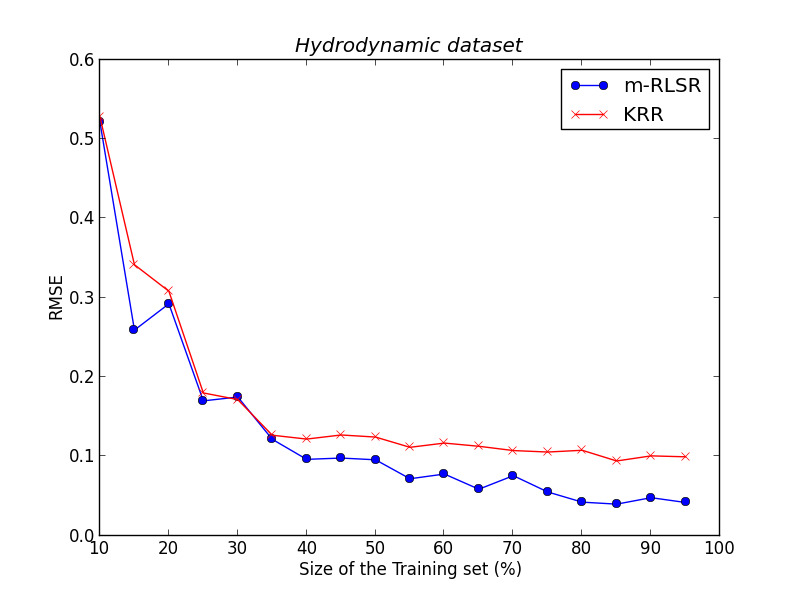}
\includegraphics[scale=0.25]{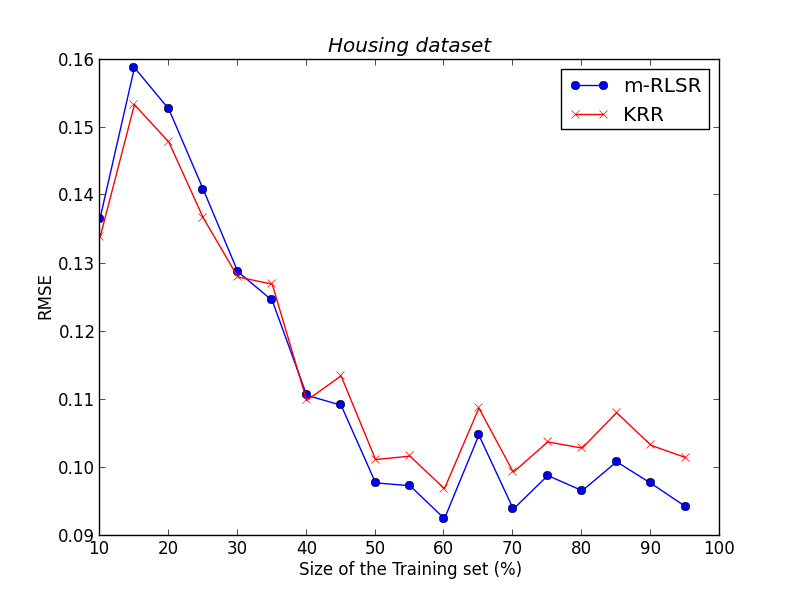}
\includegraphics[scale=0.25]{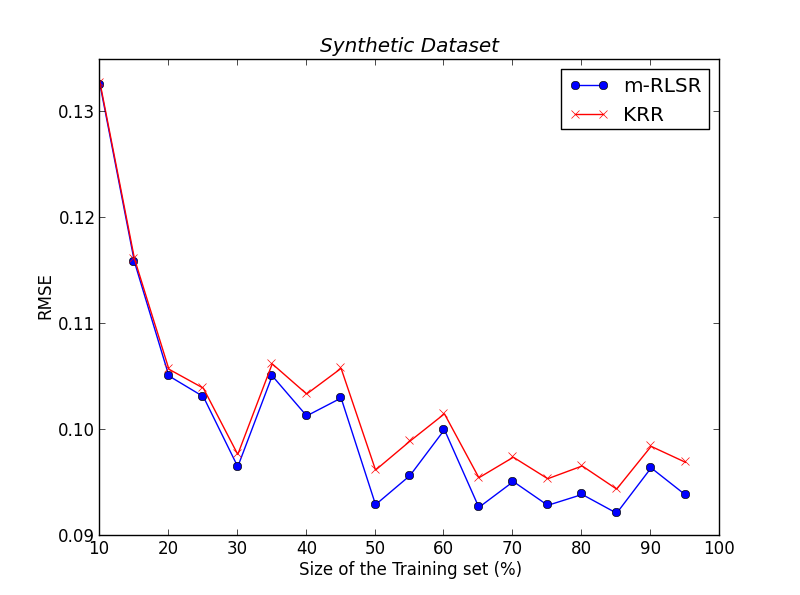}
\caption{ RMSE curve of M-RLSR (blue) and KRR (red) algorithms as a function of the dataset size. (top left) Concrete compressive strength ($m=0.1$). (top right) Yacht Hydrodynamics~($m=0.5$). (bottom left) Housing $(m=0.4)$. (bottom right) Synthetic ($m=0.1$).} 
\label{fig conv}
\end{figure}

Figure~\ref{fig conv} shows the mean of RMSE over ten run for the four datasets with \mbox{M-RLSR} and KRR when varying the number of examples of training data from 10\% to 100\% with a step size of 5\%. In this figure, we can see that \mbox{M-RLSR} with $m<1$ can improve the speed of convergence of KRR. This confirms the theoretical expectation for this situation~\cite{Mendelson10}, that is  a regularization exponent that grows significantly slower than the standard quadratic growth in the RKHS norm can lead  to better convergence behavior.


\subsubsection{Prediction Accuracy }\label{subs xp choosing lambda}

We evaluate the prediction accuracy of the M-RLSR algorithm using the datasets described above and compare it to KRR.
For each dataset we proceed as follows:  the dataset is split randomly into two parts (70\% for training and  30\% for testing), we set $\lambda=1$, and we select $m$ using cross-validation in a grid varying from  $0.1$ to $2.9$ with a step-size of $0.1$. The value of $m$ with the least mean RMSE over ten run is selected.Then, with $m$ now fixed, $\lambda$ is chosen by a ten-fold cross validation in a logarithmic grid of $7$ values, ranging from $10^{-5}$ to $10^2$.
Likewise, $\lambda_2$ for KRR is chosen by 10-fold cross-validation on a larger logarithmic grid of 25 equally spaced values between $10^{-7}$ and $10^3$.

RMSE and standard deviation (STD) results for M-RLSR and KRR are reported in Table~\ref{tab_lambda}. It is important to note that the double cross-validation on $m$ and $\lambda$  for M-RLSR,  and the cross-validation on the greater grid for the KRR takes a similar amount of  time. Table~\ref{tab_lambda} shows that the $m$-power RLSR algorithm is capable of achieving a good performance results when $m<2$.
Note that the difference between the performance of the two algorithms \mbox{M-RLSR} and KRR decreases  as the grid of $\lambda$ becomes larger, but in practice we are limited by computational reasons.

%
%
%

\begin{table}
\small
\begin{center}

\caption{Performance (RMSE and STD) of  $m$-power RLSR~(\mbox{M-RLSR}) and  KRR algorithms on synthetic and UCI datasets. m is chosen by cross-validation on a grid ranging from 0.1 to 2.9 with a step-size of $0.1$. }
\vspace{0.3cm}

\begin{tabular}{|l|cc|ccc|}
\cline{2-6}
  \multicolumn{1}{c|}{}  & \multicolumn{2}{c|}{KRR} & \multicolumn{3}{c|}{M-RLSR} \\
 \hline
 Dataset & RMSE & STD & $m$ & RMSE & STD\\
\hline

Compressive&8.04e-2&3.00e-3&1.6&7.31e-2&3.67e-3\\

Slump&3.60e-2&5.62e-3&1.1&3.52e-2&6.49e-3\\

Yacht Hydro&0.165&1.13e-2&0.1&1.56e-2&7.53e-3\\

Wine&8.65e-2&6.18e-3&1.3&8.17e-2&6.07e-3\\

Energy&4.12e-2&1.79e-3&1.1&3.79e-2&2.87e-3\\

Housing&10.6e-2&7.98e-3&1.3&7.26e-2&9.92e-3\\

Parkinson&8.05e-2&4.51e-3&0.3&5.56e-2&3.29e-3\\

Synthetic&3.19e-2&1.56e-3&0.4&1.26e-2&5.85e-4\\

\hline

\end{tabular}
\label{tab_lambda}
\end{center}

\end{table}

\end{document}